# Wind Speed Forecasting Based on Data Decomposition and Deep Learning Models: A Case Study of a Wind Farm in Saudi Arabia

Yasmeen Aldossary *⁽ᴵᴰ⁾, Nabil Hewahi **‡⁽ᴵᴰ⁾, Abdulla Alasaadi ***⁽ᴵᴰ⁾

* College of Science, University of Bahrain, Bahrain

** Department of Computer Science, College of Information Technology, University of Bahrain, Bahrain

(202100018@stu.uob.edu.bh, nhewahi@uob.edu.bh, aalasaadi@uob.edu.bh)

‡Corresponding Author; Yasmeen Aldossary, Saudi Arabia, Tel: +96 654 980 9601,

202100018@stu.uob.edu.bh

**Abstract-** With industrial and technological development and the increasing demand for electric power, wind energy has gradually become the fastest-growing and most environmentally friendly new energy source. Nevertheless, wind power generation is always accompanied by uncertainty due to the wind speed's volatility. Wind speed forecasting (WSF) is essential for power grids' dispatch, stability, and controllability, and its accuracy is crucial to effectively using wind resources. Therefore, this study proposes a novel WSF framework for stationary data based on a hybrid decomposition method and the Bidirectional Long Short-term Memory (BiLSTM) to achieve high forecasting accuracy for the Dumat Al-Jandal wind farm in Al-Jouf, Saudi Arabia. The hybrid decomposition method combines the Wavelet Packet Decomposition (WPD) and the Seasonal Adjustment Method (SAM). The SAM method eliminates the seasonal component of the decomposed subseries generated by WPD to reduce forecasting complexity. The BiLSTM is applied to forecast all the deseasonalized decomposed subseries. Five years of hourly wind speed observations acquired from a location in the Al-Jouf region were used to prove the effectiveness of the proposed model. The comparative experimental results, including 27 other models, demonstrated the proposed model's superiority in single and multiple WSF with an overall average mean absolute error of 0.176549, root mean square error of 0.247069, and R-squared error of 0.985987.

## 1. Introduction

The energy requirement is significantly growing with the rapid advancement of the times and the continuous development of science and technology. The world's dominant energy is still conventional fossil fuels; in 2021, fossil fuels accounted for 82% of primary energy [1]. Fossil fuel combustion is a primary trigger for global issues. Therefore, the development and utilization of new energy sources, which replace conventional energy sources, is essential for the sustained growth of humanity. As a new energy source, wind energy has evolved into the most promising, environment-friendly, and fastest-growing [2].

With the social and economic developments and the rapidly growing population in Saudi Arabia, the energy demand has recently increased yearly. In 2021, electricity generation resources were as follows: natural gas 60.544%, oil 39.2316%, and renewables 0.0022%, demonstrating the heavy dependency on petroleum products, which will affect Saudi Arabia's ability to export fossil fuels by cutting the country's primary source of revenue [1], [3]. As part of Saudi Arabia's 2030 Vision, the National Renewable Energy Program (NREP) was formed to increase Saudi Arabia's share of renewable energy generation, balance local energy sources, and meet the Kingdom's commitments to reduce carbon dioxide emissions [4]. In 2018 under the NREP, Saudi Arabia's northern region Al-Jouf was selected to host the Kingdom's first onshore renewable energy project dedicated to wind energy and the largest in the Middle East. The Dumat Al-Jandal 400 MW wind farm produces enough energy to power 70,000 homes, offsetting 994 thousand tons of $CO_2$ emissions per year and displacing over 894 thousand barrels of oil equivalent per year [5].

In the wind forecasting (WF) field, wind energy can be either directly forecasted based on the wind turbine output or indirectly forecasted using wind speed forecasting (WSF), which is converted into electricity. Indirect forecasting is more accurate than direct energy forecasting due to the greater spatial correlation of the wind.

According to Shumway and Stoffer [6], some wind speeds could be classified as stationary due to wind stationarity behavior, which is an important characteristic. The time series will be stationary if it satisfies three conditions: the time series' mean, variance, and covariance are time-invariant, meaning that the statistical characteristics are independent of the time, whereas the time series will be considered as non-stationary if the statistical characteristics are dependent on the time [7], [8].

Accurate wind forecasting has been associated with great significance in positively exploiting renewable energy sources. Brown et al. [9] developed a straightforward wind speed approach based on the wind power curve, taking into account the characteristics of wind speed data. Since then, in forecasting the wind speed field, a significant amount of research has been carried out, which led to the development of various methods as well as reliable and efficient technologies with different success rates across multiple wind farms [10].

WSF can be classified according to two main categories [11], one is based on the time horizons of the forecast, and the other is based on the forecasting methodology. Regarding time horizon, forecasting can be divided into very short-term (A few minutes- less than an hour) [12], short-term (One hour – less than 72 hours) [13], medium-term (Three days - one month) [14], and long-term (More than one month) [15]. On the other hand, there are mainly four WSF approaches Artificial Intelligence (AI)-based [16], Statistical [17], Physical [18], and hybrid methods [19].

Nevertheless, the wind power system is unfavorably influenced by inconsistent wind energy generation caused by the wind speeds' randomness, intermittency, and volatility. Thus, accurate wind speed forecasting will dramatically improve the wind power systems' stability, economy, reliability, security, and efficiency by reducing these unfavorable influences caused by wind speed characteristics.

In order to help the Dumat Al-Jandal wind farm to avoid the energy limitations problem, improve their wind farm system, and obtain the maximum benefit of their renewable energy source. This paper designs and develops a wind speed forecasting framework for stationary data based on data decomposition as a pre-learning stage, followed by deep learning (DL) methods aiming to achieve a high forecasting accuracy for the Dumat Al-Jandal wind farm in Al-Jouf, Saudi Arabia.

2. Literature Review

In WF, many papers have been published based on several methods to improve the WSF accuracy, such as data decomposition, including Empirical Mode Decomposition (EMD)-based, wavelet-based, and seasonal decomposition-based.

Regarding the EMD-based, Mi et al. [20] combined the EMD with the Convolutional Support Vector Machine (CNN-SVM) model and obtained significantly better performance than the benchmark models. Meanwhile, Li et al. [21] combined the EMD method with an Autoregressive Integrated Moving Average (ARIMA), a Bat Optimization, a Backpropagation Neural Network (BPNN), an Elman Neural Network (ENN), and a modified Support Vector Regression (SVR). The results showed that the developed combined model outperformed other benchmark models. In another study, Qian et al. [22] constructed a hybrid method by coupling the Complete Ensemble EMD with Adaptive Noise (CEEMDAN) with the Temporal Convolutional Networks (TCN) methods. The proposed method was examined from one to four steps ahead, and the forecasting error was reduced by nearly half. On the other hand, Ren et al. [23] investigated the performance of EMD-based along with SVR and Artificial Neural Network (ANN) models. The researchers claimed that the EMD method and its improved versions enhanced the performance of SVR significantly but marginally on ANN.

As for the wavelet method, Berrezzek et al. [24] study found that Discrete Wavelet Transform (DWT) method with Daubechies 4 wavelet (db4) and a five-level decomposition integrated with the ANN model exhibited high precision and accuracy for one-day ahead. In contrast, Saoud et al. [25] discovered that a combination of Stationary Wavelet Transform (SWT) with Quaternion-valued Neural Networks (QVNN) methods reduced the Mean Absolute Error (MAE) and Root Mean Square Error (RMSE) by 26.5% and 33%, respectively. Additionally, Jaseena and Kovoor [26] substantiated that wavelet-based and EMD-based methods with the Bidirectional Long Short-term Memory (BiLSTM) model yield better accuracy and stability than standalone models in the short-term forecast.

In order to further improve the performance of WSF, the decomposition techniques were hybridized in multiple studies, such as Duan et al. [27] hybridized the Ensemble EMD (EEMD) and Wavelet Packet Decomposition (WPD) methods with a Group Method of Data Handling (GMDH) network model and concluded that the proposed hybrid model achieved the best accuracy among all the involved models. Meanwhile, Mi et al. [28] hybridized the WPD and CEEMDAN with three ANN models, finding that the hybridized models have better prediction performance than the individual ANNs models.

Regarding the seasonal decomposition-based, Jiang et al. [29] ascertained that combining SVR, Seasonal Adjustment Method (SAM), and Elman Recurrent Neural Network (ERNN) methods have greater accuracy than the individual models. In contrast, the SAM method has proven its superiority in forecasting one to three steps by merging with an Exponential Smoothing Method (ESM) and a Radial Basis Function Neural Network (RBFNN) model in the study of Kong et al. [30].

Taking into account the recent growth of renewable energy resources in Saudi Arabia, particularly the wind resource. As a result, a small body of research has been carried out over the last decade in the WSF of several Saudi Arabia cities. Regarding the forecasting methods, most of the studies have used classical machine learning (ML) methods such as Random Forest (RF) [31], [32], and SVM [33], a few used ANN [34]–[36], and one used the Long Short-term Memory (LSTM) model [37].

*2.1. Summary*

After several investigations on the previous literature, the aforementioned forecasting approaches have certain intrinsic shortcomings. As for the individual model limitations, they generally cannot provide a reliable WSF. The hybrid Artificial Intelligence (AI) models have some limitations, such as the prolonged time consumption of training operations and the complex calculation [38].

Aside from the shortcomings of each data decomposition method, the researcher detected an important drawback. Most literature developed a decomposition-based hybrid model that tends to decompose the whole time series into subseries first. Then each subseries is split into training/test datasets, which leads to a significant ML problem denoted as data leakage [8], causing misleading and over-estimated performances than what is realistically achievable [39].

Furthermore, in most of the literature, the data decomposition methods were applied to non-stationary data, while the rest did not specify whether their data were stationary. In this respect, research in the wind forecasting field that investigates the impact of decomposing data with a stationary nature on the model performance is definitely needed.

It can be concluded that the hybrid decomposition strategies proved their effectiveness. The hybrid decomposition algorithm-based models generally perform better in accuracy and stability. However, it can be noted that different techniques were used on specific location wind speeds with different time-scale forecasting in each paper. Anyhow, determining the most appropriate combination of hybrid methods that suits all situations is laborious and nearly impossible due to the different nature of the winds in each location.

Therefore, further examination is required to search for more advanced and optimal combinations for the case study. In order to handle the challenges of the previous literature, a novel composite framework that combined an evolutionary decomposition technique with a deep learning model was proposed for short-term multi-step ahead WSF and appropriate for data with a stationary nature.

*2.2. The Study Contributions*

In this study, three contributions pertaining to WSF are reported.

The first contribution pertains to wind speed research in Saudi Arabia. The existing literature on wind forecasting in Saudi Arabia is limited. Regarding the forecasting methods, as mentioned earlier, most of the studies used classical ML methods, a few used ANN, and one used the LSTM model; this illustrates the need to investigate additional methods for forecasting wind speed in Saudi Arabia. Additionally, the current research is the first to investigate the wind speed data of the Al-Jouf region.

The second contribution is about stationary wind speed and decomposition research. Recently, existing literature has used the data decomposition method specifically for wind that is non-stationary to improve WSF accuracy. Even though the wind nature could be stationary or non-stationary, research that applies the data decomposition methods on a stationary wind for improving wind forecasting accuracy is almost nonexistent. In this study, the state of the knowledge gap will be addressed by a comprehensive investigation studying the effect of decomposition methods on stationary wind speed.

Regarding the third contribution, this study proposes an innovative hybrid approach for WSF that significantly improves forecasting accuracy. This will greatly contribute to the scientific community in their state of knowledge. The hybrid approach utilizes a novel secondary decomposition algorithm to improve the effectiveness of the BiLSTM algorithm; the novel secondary decomposition algorithm is presented using the WPD method and SAM.

**3. Methods and Procedures**

This section introduces the proposed model's methods, illustrates the proposed framework details, presents detailed information about the case study, and shows the employed evaluation metrics.

*3.1. Related Methodology*

In this study, after many experiments, the BiLSTM networks with a single hidden layer exhibited a reliable performance in solving Al-Jouf wind speed data problems and consuming less computational time than two constructed deep learning models, a staked LSTM with two hidden layers and a TCN with two TCN layers. The walk-forward technique is utilized in this study to allow the proposed hybrid model to respond to wind speed changes adaptively [39].

WPD and SAM data decomposition methods were selected based on their good performance in previous studies for reducing forecasting errors such as [28], [29].

WPD is an extended version of the DWT method [40]. WPD helped with the DWT shortage of fixed-time frequency decomposition by decomposing complex signals into different frequency bands, which will reduce the signal's noise and obtain different classification performances [41]. In a DWT process, only the previous approximation coefficients will be decomposed into a group of approximate and detail coefficients. On the other hand, the WPD will decompose both the previous approximation and detail coefficients [42]. More details on the theoretical part of the WPD method can be found in [43] and [44].

SAM is built in a way that the assumption is that seasonality can be separated from other components of the time series. The seasonal and trend components coexist in many nonlinear systems, mainly in wind speed series [45].

Multiplicative or additive operations can construct seasonal and trend components. The additive decomposition operation is more suitable for time series with seasonal variation relatively consistent with the trend [46]. For more details, refer to [47], [48].

BiLSTM is an improved version of standard LSTMs that considers past and future states to enhance prediction accuracy. In contrast, the standard LSTM only considers past observations [49], meaning that additional features can be extracted through the BiLSTM model and perform more effectively than a standard LSTM [50]. For additional information, it can be referred to [51].

*3.2. The Proposed Model Framework*

The structure of the proposed WPD-SAM-BiLSTM model is portrayed in Fig. 1, which can be described as follows,

(1) The pre-processed wind speed data are verified whether the data are stationary according to Augmented Dickey-Fuller (ADF) unit root test [52]. The data is divided into 70% for training and 30% for testing. Afterward, the data is normalized using the Z-score method.

(2) The WPD is employed to decompose the training and test signals using the Daubechies wavelet function (db22) into several subsequences with different frequency sub-bands. The WPD's components are reconstructed individually for the preparation purposes of the secondary decomposition.

(3) Seasonal component removal. Each reconstructed subseries derived from WPD is deseasonalized by the SAM method. Additionally, the seasonal indices are calculated.

(4) All the deseasonalized WPD subsequences are forecasted through the BiLSTM with a single hidden layer, then the deseasonalized WPD-BiLSTM forecasted values are adjusted by adding seasonal indices to get the final WPD-SAM-BiLSTM forecasted subseries.

(5) In order to obtain the final hybrid forecasting result, all of the WPD-SAM-BiLSTM forecasted subseries results are aggregated.

(6) The effectiveness of the proposed approach is validated through a comparison with the benchmark models.

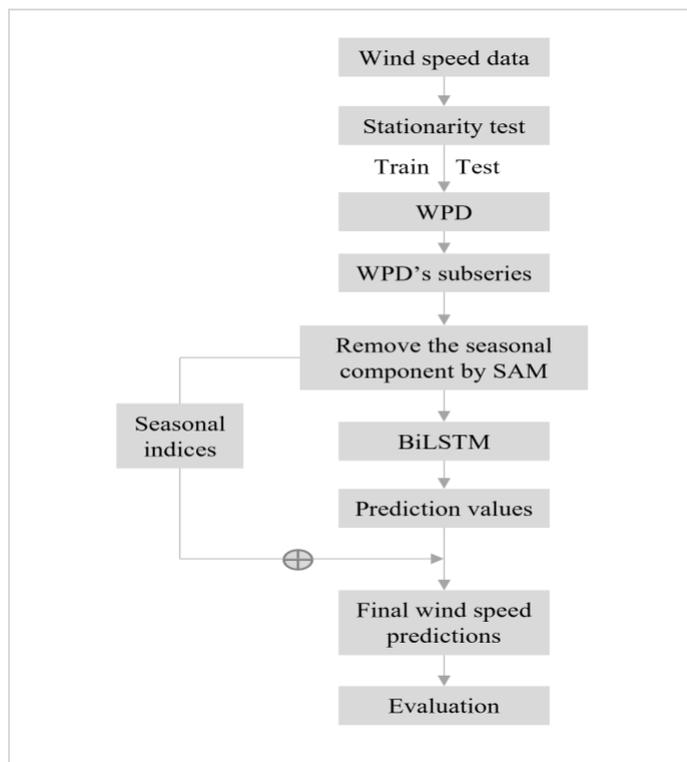

**Fig. 1.** The flow chart of the proposed WPD-SAM-BiLSTM model

*3.3. Case Study: Dumat Al-Jandal Wind Farm*

The current study case is the Dumat Al-Jandal wind farm located 900 kilometers from Riyadh, the capital of Saudi Arabia, in Dumat Al-Jandal city of the Al-Jouf region, with a latitude of 29.56° and longitude of 40.12° [5]. The wind speed data for the Al-Jouf region can be found in an official public database, namely, King Abdullah Petroleum Studies and Research Center (KAPSARC) [53]. The data consisted of the historical hourly wind speed measurement from the 1st of January 2018 to the 30th of December 2022.

As for the data preparation, first, the type of the date attribute is converted from object to date time. The data are resampled on an hourly basis. An invalid wind speed value is found, which was 999.9, and identified as a missing value. As for data outliers, any outliers greater than three standard deviations from the mean are eliminated.

Regarding the missing values, 4.27% of the Al-Jouf data were identified as missing data with 1,873 observations, and it has been treated using the forward and backward filling technique. The final data after the pre-processing consists of 43,774 wind speed observations.

*3.4. The Benchmark Models*

In order to demonstrate the effectiveness of the proposed framework in this study, 27 benchmark models were constructed. The benchmark models consist of three deep learning models LSTM, BiLSTM, and TCN or their combination with different data decomposition methods such as EMD-based, wavelets-based, and seasonal decomposition-based methods. The benchmark models are listed in Table 1.

**Table 1:** List of models used in this study as a benchmark

| Category | Abbreviation | | |
|---|---|---|---|
| Deep learning | LSTM | BiLSTM | TCN |
| Wavelet-based | WPD-LSTM | WPD-BiLSTM | WPD-TCN |
| | DWT-LSTM | DWT-BiLSTM | DWT-TCN |
| | SWT-LSTM | SWT-BiLSTM | SWT-TCN |
| EMD-based | EMD-LSTM | EMD-BiLSTM | EMD-TCN |
| | EEMD-LSTM | EEMD-BiLSTM | EEMD-TCN |
| | CEEMDAN-LSTM | CEEMDAN-BiLSTM | CEEMDAN-TCN |
| Seasonal decomposition-based | SAM-LSTM | SAM-BiLSTM | SAM-TCN |
| | STL-LSTM | STL-BiLSTM | STL-TCN |

*3.5. Performance Evaluation*

This study employs three performance measures to evaluate the proposed model forecasting accuracy: MAE, RMSE, and R-square ($R^2$), for the different forecast time horizons (i.e., 1-step, 3-step, and 5-step ahead). The RMSE is usually employed to represent the dispersion of the results, while the MAE can demonstrate the deviation of the prediction. $R^2$ is used to determine the best forecasting model with the highest accuracy based on the correlation level between predicted and actual values. The smaller the RMSE and MAE values, the more accurate the forecasting model is, whereas an $R^2$ closer to one means the more accurate forecasting model [54], [55]. The three error measures expression are given by,

$$MAE = \frac{1}{N}\sum_{t=1}^{N}|y_t - \hat{y}_t| \quad (1)$$

$$RMSE = \sqrt{\frac{1}{N}\sum_{t=1}^{N}(y_t - \hat{y}_t)^2} \quad (2)$$

$$R^2 = 1 - \frac{\sum_{t=1}^{N}(y_t - \hat{y}_t)^2}{\sum_{t=1}^{N}(y_t - \bar{y})^2}, \text{ for } \bar{y} = \frac{1}{N}\sum_{t=1}^{N}y_t \quad (3)$$

where $y_t$ and $\hat{y}_t$ are the actual and the forecasted value of a time t, respectively; $\bar{y}$ is the average of the actual values; N is the sample size.

## 4. Results and Discussion

The results of this study are given in the first section, followed by a comprehensive discussion.

*4.1. Results*

In this study, two experiments are performed. The first experiment illustrates the superiority and accuracy of the proposed WPD-SAM–BiLSTM model over the WPD-based and SAM-based hybrid models. The second experiment shows further proof of the superiority of the proposed model by providing an extra comparison between the proposed model and other hybridized models. All the experiment phases are implemented in Python using an HP device with an Intel Core i5 and 16 GB of RAM.

The selection of the best hyperparameters directly impacts the final experimental results. The detailed configuration of the BiLSTM model of the proposed model is summarized in Table 2. Furthermore, the proposed model was tested over different forecasting windows, WPD levels, and seasonal periods. A window size of 3, 9, and 15 for one, three, and five hours of forecasting, respectively, a three WPD level, and a seasonal period of 2192h produced the best forecasting accuracy for the proposed model; therefore, this study selected it as the proposed model's optimal configuration.

Regarding the stationarity test, the results of the ADF test are displayed in Table 3. The p-value for the wind speed of Al-Jouf data was 0.00, which was lower than the 0.05 significant value. That indicates the ADF unit root test null hypothesis was rejected even at the 1% significance level. In addition, the t-statistic of the ADF test was -22.99, which was less than the critical values at 1%, 5%, and 10% significance levels, which was another evidence for rejecting the null hypothesis and considering the Al-Jouf time series data as stationary.

**Table 2.** Parameter setting of WPD-SAM-BiLSTM model

| Parameters | Definition | Optimal | Tested values | Selection method |
|---|---|---|---|---|
| Hidden layers | Number of hidden layers | 1 | 1, 2, and 3 | Trial/Test |
| epochs | The total number of iterations of all the training data in one cycle for training the model | 100 | 10, 100, 200, 500, 1000, and 2000 | |
| batch_size | The number of complete passes through the training dataset | 512 | 32, 64, 128, 256, 512, and 1024 | |
| Hidden units | Number of hidden units | 64 | 16, 32, 64, 128, and 256 | |
| Dropout | A mechanism to improve the generalization of neural nets | 0.0 | 0.0, 0.01, 0.1, and 0.2 | Bayesian optimization |
| loss | A function that compares the target and predicted output values | mae | mae and mse | |
| learning_rate | A tuning parameter in an optimization algorithm that determines the step size at each iteration while moving toward a minimum of a loss function | 0.0001 | 0.01, 0.001, 0.0001, and 0.0075 | |

**Table 3.** The test results of ADF unit root for Al-Jouf data

| Variable | Definition | Value |
|---|---|---|
| t-statistic for ADF | The ratio of the departure of the estimated value of a parameter from its hypothesized value to its standard error. | -22.991492 |
| p-value | The probability under the assumption of no effect or no difference (null hypothesis), of obtaining a result equal to or more extreme than what was actually observed. | 0.0000 |
| Test critical values | a point on the distribution of the test statistic under the null hypothesis that defines a set of values that call for rejecting the null hypothesis. | **1% level**: -3.430499 **5% level**: -2.861606 **10% level**: -2.566805 |

*4.1.1. Experiment I*

In this experiment, first, the performance of non-decomposed deep learning models and the decomposed ones using the WPD and SAM methods were compared to emphasize the importance of the WPD data decomposition approach and to identify the advantage of deseasonalizing using the SAM decomposition method on the subjected data. Table 4 compares the deep learning models LSTM, BiLSTM, and TCN performance in forecasting the data with and without the WPD and SAM methods for one, three, and five hours ahead based on MAE, RMSE, and $R^2$.

Regarding the WPD method, it can be observed that the deep learning models with WPD perform better than the non-decomposed ones. Especially in LSTM, the decomposed data has significantly reduced MAE values with 0.550397, 0.678599, and 0.556770 for forecasting one, three, and five hours, respectively. Similarly, in the evaluation error RMSE and $R^2$, the deep learning models have been affected positively with the WPD decomposed data. It can be concluded that the best resulting WPD-based model was the WPD-LSTM.

As for the SAM method, the MAE was improved in all deep learning models; the MAE has been reduced with ranges from 0.17 to 0.19, 0.23 to 0.25, and 0.27 to 0.33 in one-step, three-step, and five-step forecasting, respectively. At the same time, in RMSE, the reduction ranged from 0.28 to 0.30, 0.34 to 0.37, and 0.39 to 0.46 for one, three, and five hours of forecasting, respectively. Meanwhile, $R^2$ showed an increase in deep learning with the SAM method; the improvement ranged from 0.14 to 0.15, 0.20 to 0.22, and 0.25 to 0.30 for 1-step, 3-step, and 5-step ahead, respectively. Furthermore, according to the forecasting accuracy, the SAM-LSTM was identified as the best SAM-based model.

Figure 2 compares the performance of the proposed model with the WPD-LSTM and SAM-LSTM in forecasting AL-Jouf wind speed data to demonstrate the superiority of the hybrid data decomposition method at all the forecasting time horizons.

Compared with the best resulting models from Table 3, it is clear from Fig. 2 that the proposed model achieved the lowest MAE and RMSE and the highest R-squared. To be more specific, the proposed model delivered an outstanding forecasting accuracy with minimum MAE values of 0.116901, 0.155379, and 0.257366 in one, three, and five hours ahead, respectively, and with minimum RMSE values of 0.152297, 0.223525, and 0.365386 for one, three, and five hours forecasting, respectively. The model also attained maximum $R^2$ values of 0.995279, 0.989834, and 0.972849 in 1-step, 3-step, and 5-step ahead forecasting, respectively.

Therefore, the proposed model has proved its superiority over the individual DL, hybrid WPD-based, and SAM-based models because it delivered better forecasting accuracy than the WPD-LSTM and SAM-LSTM models in all tested time horizons.

**Table 4.** The forecasting errors of deep learning, WPD-based, and SAM-based models

| Model | One hour | | | Three hours | | | Five hours | | |
|---|---|---|---|---|---|---|---|---|---|
| | MAE | RMSE | $R^2$ | MAE | RMSE | $R^2$ | MAE | RMSE | $R^2$ |
| LSTM | 0.997061 | 1.355951 | 0.625834 | 1.197024 | 1.608870 | 0.473428 | 1.354972 | 1.809059 | 0.334567 |
| BiLSTM | 0.994108 | 1.364204 | 0.621265 | 1.197861 | 1.601144 | 0.478473 | 1.320021 | 1.762494 | 0.368382 |
| TCN | 1.003210 | 1.382427 | 0.611079 | 1.210578 | 1.628810 | 0.460295 | 1.337749 | 1.784077 | 0.352817 |
| WPD-LSTM | 0.446663 | 0.587468 | 0.929766 | 0.518424 | 0.816262 | 0.864459 | 0.798201 | 1.238553 | 0.688095 |
| WPD-BiLSTM | 0.449233 | 0.594439 | 0.928089 | 0.734356 | 1.075251 | 0.764802 | 1.029353 | 1.472736 | 0.558996 |
| WPD-TCN | 0.753671 | 0.998115 | 0.797261 | 1.107850 | 1.504633 | 0.539451 | 1.263541 | 1.701495 | 0.411350 |
| SAM-LSTM | 0.813669 | 1.068217 | 0.767782 | 0.949024 | 1.241478 | 0.686459 | 1.033784 | 1.350201 | 0.629322 |
| SAM-BiLSTM | 0.812761 | 1.068655 | 0.767591 | 0.950915 | 1.242309 | 0.686039 | 1.036215 | 1.350134 | 0.629358 |
| SAM-TCN | 0.830729 | 1.094498 | 0.756215 | 0.976959 | 1.279423 | 0.667000 | 1.066070 | 1.386675 | 0.609025 |

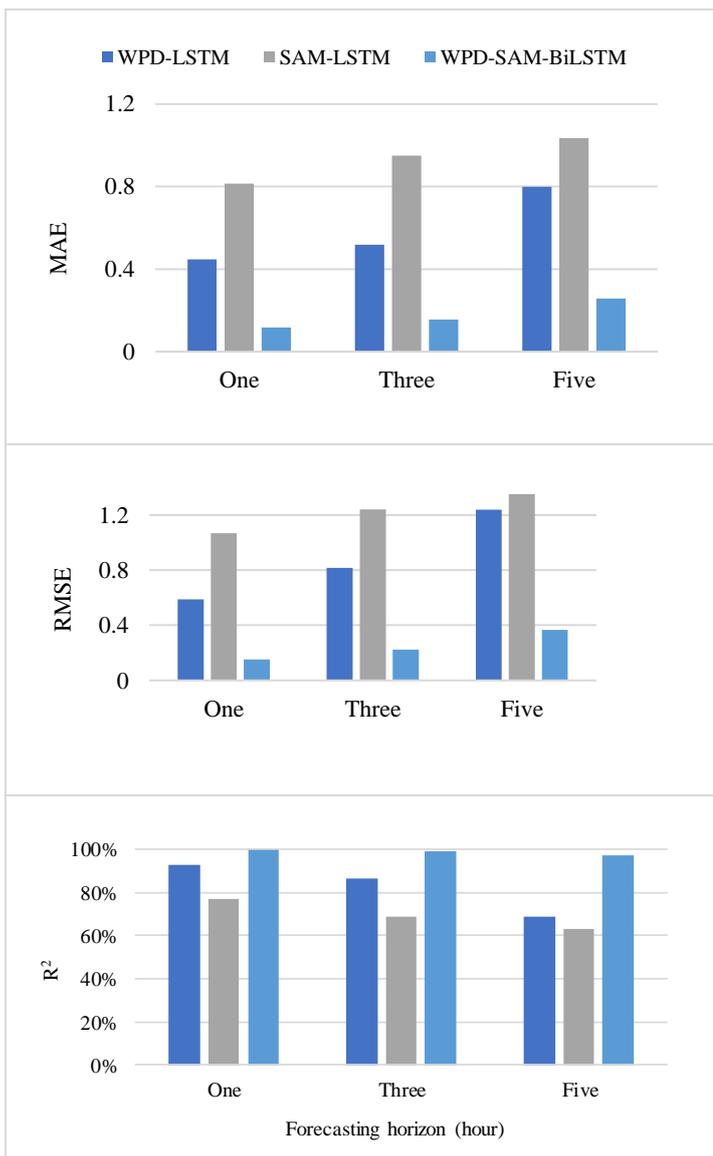

**Fig. 2.** The forecasting errors of the proposed model, WPD-LSTM, and SAM-LSTM

### 4.1.2. Experiment II

This experiment was conducted to illustrate further the proposed model's superiority against the benchmark models: the hybridization of the LSTM, BiLSTM, and TCN models with other data decomposition methods, i.e., EMD, EEMD, CEEMDAN, DWT, SWT, and Seasonal-Trend decomposition based on Loess (STL). The performance metrics in forecasting one and multi-step, including MAE, RMSE, and $R^2$ of the benchmark models and the proposed model, are displayed in Table 5. The results of Table 5 can be outlined as follows:

(1) The proposed model's forecasting errors had the lowest MAE and RMSE and the highest $R^2$ compared to all the benchmark models in one, three, and five hours of forecasting. Therefore, the proposed WPD-SAM-BiLSTM model based on the hybrid decomposition technique has the best performance in forecasting the case study data.

(2) In the EMD-based models, combining the EEMD method and LSTM model produced the minimum MAE and RMSE and the maximum $R^2$ for all time forecasting horizons. In contrast, the combination of the EMD method and TCN model had the worst performance.

(3) Regarding the wavelet-based models, the SWT-BiLSTM performed better than the other models for all evaluation errors in one-hour forecasting. Meanwhile, the performance of the SWT-LSTM model was the best in the three and five hours ahead forecasting according to the evaluation errors.

(4) Among the three decomposition-based models, the STL-based models delivered the worst performance in all forecasting errors of each tested time horizon.

(5) The hybrid LSTM-based and BiLSTM-based models achieved better forecasting performance with minimum MAE and RMSE and maximum $R^2$ in one, three, and five hours of forecasting. On the contrary, the hybrid TCN-based models always delivered the worst performance in all cases.

(6) As for the forecasting time horizon, generally, in terms of MAE and RMSE, the longer the forecasting horizon, the higher the forecasting error is. In contrast, regarding the $R^2$, when the forecasting horizon increases, the $R^2$ decreases.

### 4.2. Discussion

In order to help the Dumat Al-Jandal wind farm decision-makers develop and enhance their wind farm system, such as anticipating the future trend of wind energy, mitigating the negative influences on electrical grids, fulfilling the energy demand, and minimizing the financial losses, this study has designed and developed a comprehensive multi-step wind speed forecasting model based on a deep learning BiLSTM model and a WPD-SAM hybrid data decomposition method for forecasting Al-Jouf data.

#### 4.2.1. The effect of data decomposition on stationary data

This study investigated various decomposition methods to fill a gap in knowledge concerning the data decomposition methods' effectiveness on stationary data. Mainly, three different decompositions were applied: EMD-based, wavelets-based, and seasonal decomposition-based methods and combined with LSTM, BiLSTM, and TCN models. The following discussion is according to Table 4 and Table 5.

Table 5. Performance indices of the proposed model compared to other decomposition methods

| Model | One hour | | | Three hours | | | Five hours | | |
|---|---|---|---|---|---|---|---|---|---|
| | MAE | RMSE | $R^2$ | MAE | RMSE | $R^2$ | MAE | RMSE | $R^2$ |
| EMD-LSTM | 0.505301 | 0.728326 | 0.892048 | 0.648687 | 0.876421 | 0.843742 | 0.749505 | 1.005685 | 0.794353 |
| EMD-BiLSTM | 0.555937 | 0.767302 | 0.880185 | 0.716384 | 0.965154 | 0.810500 | 0.851693 | 1.133430 | 0.738791 |
| EMD-TCN | 0.829052 | 1.082326 | 0.761607 | 1.023024 | 1.340570 | 0.634410 | 1.139150 | 1.491248 | 0.547834 |
| EEMD-LSTM | 0.343763 | 0.483686 | 0.952442 | 0.498572 | 0.688129 | 0.903778 | 0.592491 | 0.808241 | 0.867195 |
| EEMD-BiLSTM | 0.376155 | 0.52123 | 0.944772 | 0.538275 | 0.734514 | 0.890369 | 0.653781 | 0.894760 | 0.837242 |
| EEMD-TCN | 0.722558 | 0.968802 | 0.809207 | 0.929027 | 1.229399 | 0.692871 | 1.033863 | 1.364306 | 0.621596 |
| CEEMDAN-LSTM | 0.467198 | 0.669778 | 0.9087068 | 0.6188102 | 0.838183 | 0.857080 | 0.738797 | 0.738797 | 0.801274 |
| CEEMDAN-BiLSTM | 0.472186 | 0.674155 | 0.907509 | 0.622495 | 0.839164 | 0.856745 | 0.741579 | 0.992848 | 0.799569 |
| CEEMDAN-TCN | 0.798767 | 1.060069 | 0.771311 | 1.000446 | 1.309495 | 0.651162 | 1.118134 | 1.461092 | 0.565936 |
| DWT-LSTM | 0.444658 | 0.586505 | 0.929996 | 0.514807 | 0.811024 | 0.866193 | 0.798701 | 1.236146 | 0.689306 |
| DWT-BiLSTM | 0.449051 | 0.593646 | 0.928281 | 0.757322 | 1.132086 | 0.739282 | 1.058601 | 1.504442 | 0.539803 |
| DWT-TCN | 0.763192 | 1.013253 | 0.791064 | 1.108947 | 1.503270 | 0.540285 | 1.266322 | 1.698731 | 0.413261 |
| SWT-LSTM | 0.357765 | 0.462391 | 0.956489 | 0.447874 | 0.698974 | 0.95400612 | 0.678798 | 1.087012 | 0.759751 |
| SWT-BiLSTM | 0.352946 | 0.458015 | 0.957308 | 0.608645 | 0.942834 | 0.819164 | 0.967772 | 1.430421 | 0.583974 |
| SWT-TCN | 0.728412 | 0.965170 | 0.810423 | 1.101392 | 1.490937 | 0.547797 | 1.259435 | 1.694049 | 0.416490 |
| STL-LSTM | 0.950841 | 1.267780 | 0.672912 | 1.059871 | 1.399590 | 0.601509 | 1.021396 | 1.337648 | 0.636180 |
| STL-BiLSTM | 0.949095 | 1.268360 | 0.672613 | 1.058139 | 1.396745 | 0.603128 | 1.010902 | 1.322288 | 0.644487 |
| STL-TCN | 0.958671 | 1.277286 | 0.667988 | 1.107543 | 1.463451 | 0.564315 | 1.175814 | 1.546917 | 0.513442 |
| **WPD-SAM-BiLSTM** | **0.116901** | **0.152297** | **0.995279** | **0.155379** | **0.223525** | **0.989834** | **0.257366** | **0.365386** | **0.972849** |

When comparing the EMD-based hybrid models with non-decomposed models, the forecasting accuracy of the former was higher than the latter significantly, especially for the EEMD-LSTM hybrid model. For example, in five hours of forecasting, after the stationary data were decomposed through the CEEMDAN method, the BiLSTM performance improved by 43.82%, 43.66%, and 117.04% in MAE, RMSE, and $R^2$, respectively. This aligns with the results of the studies conducted by Mi et al. [20], Li et al. [21], and Jaseena and Kovoor [26], in which they stated that the EMD method improved the WSF models outstandingly. Furthermore, a similar conclusion was drawn by Qian et al. [22], which was that the EMD-based hybrid methods, particularly the CEEMDAN method, can enhance the forecasting error significantly.

The forecasting accuracy has greatly improved when comparing the wavelet-based hybrid models with non-decomposed models, specifically for the SWT-LSTM and SWT-BiLSTM models. For illustration, in one step ahead forecasting, the DWT method enhanced the LSTM performance by 55.40%, 56.74%, and 48.60% in MAE, RMSE, and $R^2$, respectively. This outcome agrees with Berrezzek et al. [24], who concluded that the wavelet-based methods, particularly the DWT method, delivered more efficient performance than the non-decomposed ones. Saoud et al. [25] also concluded that SWT reduced the forecasting error by 27% and 33% for MAE and RMSE, respectively.

In the seasonal decomposition-based hybrid models, the forecasting accuracy was higher than in the non-decomposed models; even though the improvement was not that significant, the seasonal method affected the stationary data positively. For instance, in five hours of forecasting, the STL method improved the TCN performance by 4.43%, 7.61%, and 9.31% in MAE, RMSE, and $R^2$, respectively. The results support what was reported by Jiang et al. [29] and Kong et al. [30]; they found that the seasonal methods positively affect the models' performance.

Therefore, it can be concluded that the EMD-based, wavelet-based, and seasonal decomposition-based not only enhance the forecasting accuracy of the stationary data but also can effectively reduce the MAE and RMSE and increase the $R^2$, indicating that even though each method function and process the data differently, applying the data decomposition as a pre-processing data method on data with a stationary nature could improve the forecasting accuracy enormously.

*4.2.2. The superiority of the proposed model*

This study conducted several comparisons to reveal the proposed model's superiority and verify the contribution of each part of the proposed model. The effectiveness of WPD and SAM methods over the non-decomposed deep learning models was emphasized in Table 4. WPD and SAM methods made a notable contribution in improving the forecasting accuracy of the deep learning models, which means that the deep learning models cannot sufficiently describe the volatility of Al-Jouf wind speed, and decomposing the data using WPD and SAM methods helped these models in capturing more features. These findings are compatible with those reported by other studies, such as Kong et al. [30] for the SAM method and Duan et al. [27], and Mi et al. [28] for the WPD method.

The superiority of the proposed model over WPD-based and SAM-based models was verified in Fig. 2. The proposed model revealed a remarkable forecasting accuracy in single and multistep forecasting compared with WPD-based and SAM-based models, indicating that combining the WPD and SAM decomposing methods with the BiLSTM model is an effective combination in describing the Al-Jouf wind speed volatility. Regardless of the methods used, a similar conclusion was drawn by Duan et al. [27] and Mi et al. [28] that combining two decomposition methods improves the results significantly rather than a single decomposition method.

This study further proved the dominance of the proposed model over the hybridization of the EMD, EEMD, CEEMDAN, DWT, SWT, and STL decomposition methods with the DL-based models, such as LSTM, BiLSTM, and TCN. With reference to Table 5, the highest forecasting accuracy of the Al-Jouf wind speed data belongs to the proposed model over single and multistep forecasting; this suggests that the hybrid methodology WPD-SAM-BiLSTM captures better the forecaster's behavior and highlights the proposed model superiority over other hybridization methods.

## 5. Conclusion and Future Work

With rapid technological and economic advancements worldwide, there is a sharp increase in electrical energy demand, which led to the overexploitation of fossil fuels, such as natural gas, oil, and coal. The limitations of fossil fuel reserves and their environmental disruption made renewable energy, such as wind energy, an inevitable trend in the development of energy. Wind energy has evolved rapidly in recent years. Developing an advanced and accurate wind forecasting system can control the power system to preserve safe and dependable operation. However, accurate forecasting is challenging owing to wind speed characteristics such as intermittency and stochastic nature.

In order to assist the Dumat Al-Jandal wind farm in improving its system, avoiding the issue of energy limitations, and reaping the benefits of its renewable energy source to the fullest, this study developed a novel wind speed forecasting framework for stationary data based on data decomposition and deep learning methods. The developed framework was designed specifically for the Dumat Al-Jandal wind farm in Al-Jouf, Saudi Arabia, aiming to achieve high forecasting accuracy. The novel framework hybridized two decomposition methods, WPD and SAM, with a deep learning model BiLSTM; the WPD method was employed for decomposing the data into several sub-bands; the SAM method was used to deseasonalized the data; the BiLSTM was utilized to forecast the deseasonalized-decomposed subseries.

Additionally, this study contributed to the state of knowledge regarding the effect of decomposing a stationary time series on forecasting performance. In recent years, many scholars have employed the data decomposition method mainly to improve the forecast wind speed for wind that is non-stationary. Although the nature of the winds can be stationary or non-stationary, depending on their geographical location, research that utilized the data decomposition methods for improving the wind forecasting accuracy of a stationary wind is almost nonexistent. In order to fill this research gap, this study investigated the effects of that data decomposition on stationary data in terms of forecasting accuracy.

Several experiments were conducted against benchmark models over single and multiple time horizons to verify the performance of the proposed WPD-SAM-BiLSTM model. The main conclusions of this study are summarized as follows:

a) The data decomposition methods could enormously enhance the forecasting accuracy of a stationary wind.

b) The shorter the forecasting horizon, the smaller the forecasting error is.

c) The proposed model performs remarkably better than all benchmark models with an MAE of 0.116901, an RMSE of 0.152297, and an $R^2$ of 0.995279 in forecasting one hour, an MAE of 0.155379, an RMSE of 0.223525, and an $R^2$ of 0.989834 in forecasting three hours, and an MAE of 0.257366, an RMSE of 0.365386, and an $R^2$ of 0.972849 in forecasting five hours; this indicates that the proposed model is highly suitable for the stationary wind of Al-Jouf data in single and multiple forecasting.

Although the proposed forecasting model has demonstrated satisfactory forecasting accuracy results, there is still room for improvement, such as testing the proposed model on additional stationary data, other data decomposition methods, additional meteorological factors, and longer time-scale forecasting.